# AUTOMATIC FEATURE RECOGNITION AND DIMENSIONAL ATTRIBUTES EXTRACTION FROM CAD MODELS FOR HYBRID ADDITIVE-SUBTRACTIVE MANUFACTURING

Muhammad Tayyab Khan[1,3], Wenhe Feng[1], Lequn Chen[2], Ye Han Ng[2], Nicholas Yew Jin Tan[1], Seung Ki Moon[3]

[1] Singapore Institute of Manufacturing Technology (SIMTech),
Agency for Science, Technology and Research (A*STAR), 5 CleanTech Loop, #01-01 CleanTech Two
Block B, Singapore 636732, Republic of Singapore
[2] Advanced Remanufacturing and Technology Centre (ARTC),
Agency for Science, Technology and Research (A*STAR),3 CleanTech Loop, #01-01 CleanTech Two,
Singapore 637143, Republic of Singapore
[3] School of Mechanical and Aerospace Engineering, Nanyang Technological University, 639798,
Singapore

## ABSTRACT

*The integration of Computer-Aided Design (CAD), Computer-Aided Process Planning (CAPP), and Computer-Aided Manufacturing (CAM) plays a crucial role in modern manufacturing, facilitating seamless transitions from digital designs to physical products. However, a significant challenge within this integration is the Automatic Feature Recognition (AFR) of CAD models, especially in the context of hybrid manufacturing that combines subtractive and additive manufacturing processes. Traditional AFR methods, focused mainly on the identification of subtractive (machined) features including holes, fillets, chamfers, pockets, and slots, fail to recognize features pertinent to additive manufacturing. Furthermore, the traditional methods fall short in accurately extracting geometric dimensions and orientations, which are key factors for effective manufacturing process planning. This paper presents a novel approach for creating a synthetic CAD dataset that encompasses features relevant to both additive and subtractive machining through Python Open Cascade. The Hierarchical Graph Convolutional Neural Network (HGCNN) model is implemented to accurately identify the composite additive-subtractive features within the synthetic CAD dataset. The proposed model demonstrates remarkable feature recognition accuracy of around 97% and a dimension extraction accuracy of 100% for identified features. Therefore, the proposed methodology enhances the integration of CAD, CAPP, and CAM within hybrid manufacturing by providing precise feature recognition and dimension extraction. It facilitates improved manufacturing process planning by enabling more informed decision-making. The key novelty and contribution of the proposed methodology lie in its ability to recognize a wide range of manufacturing features and to precisely extract their dimensions, orientations, and stock sizes.*

## 1. INTRODUCTION

In modern manufacturing, the seamless integration of Computer-Aided Design (CAD), Computer-Aided Process Planning (CAPP), and Computer-Aided Manufacturing (CAM) is foundational. CAD facilitates the creation of detailed digital designs, marking the essential first steps in the manufacturing process [1]. Following this, CAPP plays a pivotal role by serving as the link between CAD and CAM that analyzes the necessary manufacturing processes and conditions [2]. Finally, CAM utilizes the insights provided by CAPP to control and automate the manufacturing equipment to transform a CAD design into a physical product [3]. This integrated approach ensures a smooth transition from digital blueprints to real-world products. Furthermore, this integrated method has significantly improved the efficiency of the manufacturing process. For example, Basinger et al. [2] showcased this advancement by introducing a hybrid manufacturing process planning system that automates process plans, achieving a substantial 35% reduction in machining time for a custom patient-specific bone plate.

Within the context of CAD, CAPP, and CAM's integrated workflow, one of the primary challenges faced by CAPP is automatic feature recognition (AFR). AFR is a key step in CAPP, involving the interpretation of CAD models and their translation into recognizable manufacturing features that CAM systems can use [3]. The accurate recognition of these features is crucial for generating effective manufacturing plans and optimizing the process parameters, making AFR a significant area of research in manufacturing automation.

For over four decades, researchers have been studying AFR to find different ways to detect machining features in CAD models [4,5]. Presently, AFR methodologies are primarily divided into two categories: rule-based and learning-based approaches [4–8]. For rule-based approaches, researchers must deeply understand the structures of machining features to create effective rules [8]. These rules are designed to detect machining features by analyzing their geometric and topological characteristics. In contrast, learning-based approaches employ machine learning (ML) and deep learning (DL) techniques. These techniques identify the relationships between machining features and their fundamental representations through extensive training data, moving away from the conventional dependence on predefined rules [9]. This technique has demonstrated significant promise in enhancing the accuracy and reliability of identifying complex and diverse machining features. For example, Muraleedharan and Muthuganapathy [10] utilized a learning-based approach with discrete Gauss maps and Random Forest classifiers for 24 distinct machining feature recognition. They achieved a test accuracy of 97.90% for single feature recognition with a reduced run time off 3.19 seconds. Similarly, Colligan et al. [11,12] developed the Hierarchical CADNet model based on graph convolutional neural network (GCNN) to recognize complex machining features in CAD models. They achieved a test accuracy of 97.37%, demonstrating its exceptional ability to process complex datasets with multiple intersecting machining features.

Despite remarkable advancements in learning-based AFR it still faces some challenges. For example, most existing learning-based AFR methods [4,7,9,11,13–16] predominantly focus on machining features and do not consider the vast possibilities of additive manufacturing features such as extrusions. This oversight is particularly significant in hybrid manufacturing, which integrate machining and additive manufacturing [17]. The development of an AFR architecture that comprehensively recognizes both subtractive and additive features is crucial for advancing hybrid manufacturing processes. While there are attempts to recognize both types of manufacturing features, these efforts are limited in the scope of the number of distinct features recognized. For example, Shi et al. [18] used a 2D CNN for the recognition of only 10 isolated manufacturing features, including five additive and five subtractive features.

In addition to the difficulty of recognizing a diverse range of manufacturing features, significant research gaps exist in AFR methods, particularly concerning key manufacturing process planning elements. These gaps include the need for automatic extraction of stock sizes, and geometric dimensions for each identified feature, along with their orientations. These dimensional attributes are essential for subsequent decision-making during manufacturing process planning.

To tackle these challenges, this paper proposes a novel approach to recognize hybrid additive-subtractive manufacturing features and extract the geometrical attributes from CAD models simultaneously. These attributes include feature dimensions, orientations, and stock sizes. In the proposed method, a synthetic CAD dataset is created to encompass a wide range of subtractive and additive manufacturing features, which are used to train a Hierarchical GCNN model for hybrid feature recognition. After feature recognition, each identified features in the CAD models are analyzed to extract feature dimensions, orientations, and stock sizes, utilizing the Python Open Cascade (PyOCC) [19]. The trained Hierarchical GCNN model is applied to identify and characterize features, such as dimensions, orientations, and stock sizes.

The rest of the paper is organized as follows: Section 2 outlines the framework and methodology. Section 3 focuses on the results and discussion. The paper concludes with final thoughts and directions for future research in Section 4.

## 2. METHODOLOGY

### 2.1 Overview

The proposed framework introduces a significant advancement in the CGNN model [11] by integrating new additive features, termed as extrusions. As depicted in Figure 1, the entire process is divided into two key phases: the Model Training Pipeline and the Model Inference Process. The model training pipeline initiates with the addition of new additive features to the machining feature dataset, setting the foundation for a more comprehensive and versatile model. Following this step, the process starts with the generation of hybrid additive-subtractive CAD dataset. A synthetic labeled dataset of 150,000 STEP files, encompassing both additive and subtractive features, is generated using PyOCC [19], effectively addressing the previously identified gap. This broader dataset enables the model, initially designed for machining features, to identify a more comprehensive set of hybrid features. After developing the dataset, the STEP files are converted into Boundary Representation (B-Rep) Hierarchical Graphs. These graphs are then organized into mini batches, aligning with the model's training process. Through hyperparameter optimization, the model is fine-tuned for precision in feature recognition.

The model inference process begins once the model is trained, where a user-provided STEP file is introduced for automatic feature recognition. This approach allows for the precise quantification of geometric dimensions and orientations, distinguishing the work from conventional applications of existing models.
Furthermore, the development of a unique method for calculating minimum and maximum stock sizes, during the phase represents a significant advancement in the field, offering new tools for decision-making in hybrid manufacturing processes. These methodologies, distinct from the model's initial capabilities, underscore the novelty and significance of the contributions to the field. Details of each section are explained below.

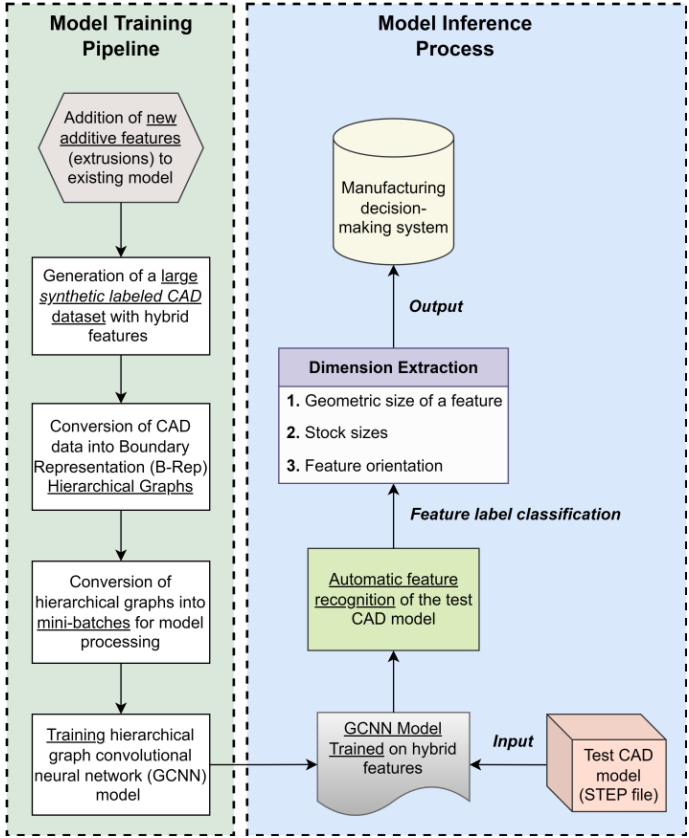

**FIGURE 1:** OVERVIEW OF THE PROPOSED FRAMEWORK

## 2.2 Dataset Development

The method initiates by integrating additional additive features into the GCNN model. These features encompass a variety of additive features such as cylindrical, rectangular/square, triangular, hexagonal, and pentagonal extrusions. This development increases the model's versatility, facilitating the recognition of 29 distinct manufacturing features. Each feature is labeled from 0 to 28. For each extrusion feature, distinct classes are established, which define their geometric and directional characteristics. This process leverages the PyOCC library to simulate material addition accurately. Additive features are differentiated from subtractive ones using predefined geometric sketches and direction vectors, a crucial step that ensures a precise representation of additive manufacturing processes. Figure 2 presents 29 distinct manufacturing features, each identified by a unique label.

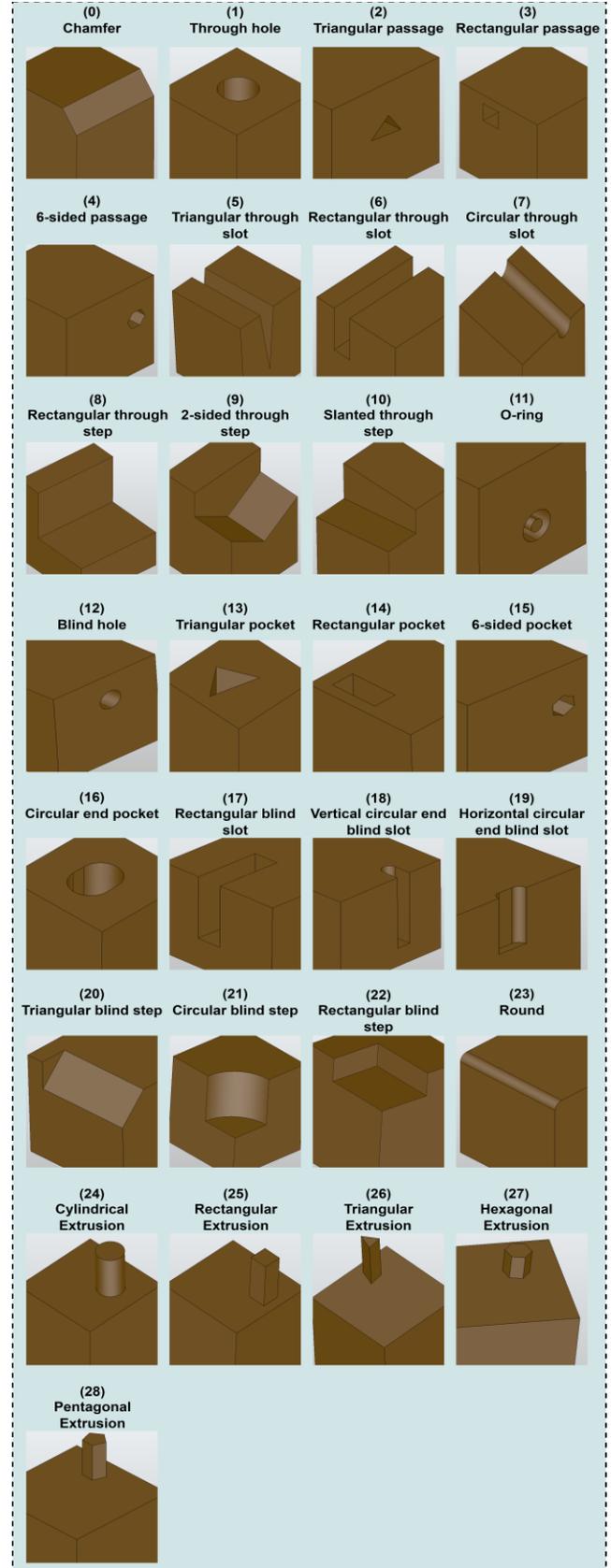

**FIGURE 2:** FEATURE CLASS WITH INDEX (LABEL)

To demonstrate the capabilities of the proposed model, we developed a PyOCC-based script for the automatic generation of 3D models with random combinations of subtractive and additive features. Utilizing the *Numba* library [20] for computational efficiency, the script generates 150,000 labeled STEP files. Each model contains 4 to 8 manufacturing features. These are now categorized into six distinct groups as:

1. Edge profiling features (round and chamfer)
2. Steps
3. Slots
4. Through features (passages or holes)
5. Blind features (pockets or holes)
6. Additive features (extrusions)

The dataset contains 30 distinct categories, including 29 manufacturing features plus the stock face. A cuboidal stock, with dimensions randomly varying between 30 to 70 arbitrary units in each axis is used. By employing python *multiprocessing* pool [21], the script efficiently produces a diverse array of models, each presenting a unique scenario of hybrid features applied to a base stock. Examples of generated labeled CAD models with hybrid features are presented in Figure 3.

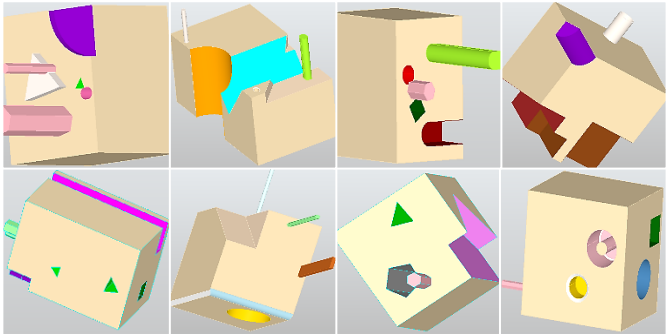

**FIGURE 3:** LABELED CAD MODELS WITH UNIQUE COLOR FOR EACH FEATURE

The B-Rep shape representation, common in CAD systems, presents challenges for direct neural network input due to its continuous data format [22]. This format includes continuous geometric and topological information that neural networks struggle to process without conversion. To address this, the synthetic labeled CAD dataset has been transformed into B-Rep Hierarchical graphs, enabling their use as discrete inputs for a GCNN model. The conversion process of CAD models into hierarchical graphs, particularly for machining features, is detailed in the paper [11]. Here, we provide a brief overview of how this method is applied to our dataset, containing both additive and machining features.

This conversion aims to capture the spatial relationships and geometric details of hybrid features, including surface geometry, edge convexity, and face topology. A two-level hierarchical graph representing the complex topology of CAD models is shown in Figure 4. The first level, known as the B-Rep face adjacency graph, outlines the model's topology. It highlights the convexity of edges between faces and attributes characteristics such as the manufacturing feature class label, face type, area, and centroid coordinates to each vertex. The second level, known as the mesh facet graph, represents the surface geometry using triangular mesh facets. Each vertex in this graph is encoded with the planar equation derived from the points of the facet and its normal vector.

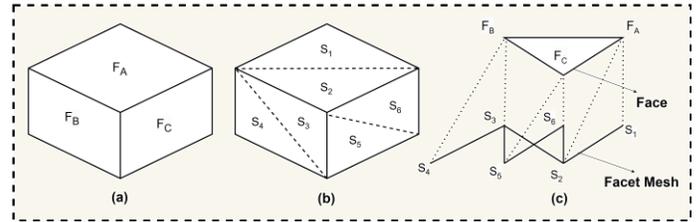

**FIGURE 4:** HIERARCHICAL GRAPH STRUCTURE (a) B-REP (b) FACET MESH (c) HIERARCHICAL GRAPH

The above two-level approach facilitates a compact yet comprehensive description of CAD models, bridging the gap between the continuous nature of B-Rep structures and the discrete input requirements of the GCNN model. The hierarchical graphs are stored in an HDF5 format [23] for efficient data handling. For neural network training, the models are batched into a new, smaller HDF5 file. This file includes a randomized graph order, normalized features and labels, and consistent input sizes based on vertex counts. This approach converts CAD data into a format suitable for machine learning. It enhances the model's capacity to process and learn from a variety of geometric configurations of hybrid features.

Upon the transformation of CAD models into hierarchical graphs, the next step involves the development of a GCNN model capable of processing these graphs. A concise summary is presented on the use of labeled CAD datasets with hybrid features in the GCNN model, whose architecture is depicted in Figure 5. The model includes spatial graph convolution layers, pooling layer, transfer layers, and residual connections. Detailed explanations of these components are in the following sections.

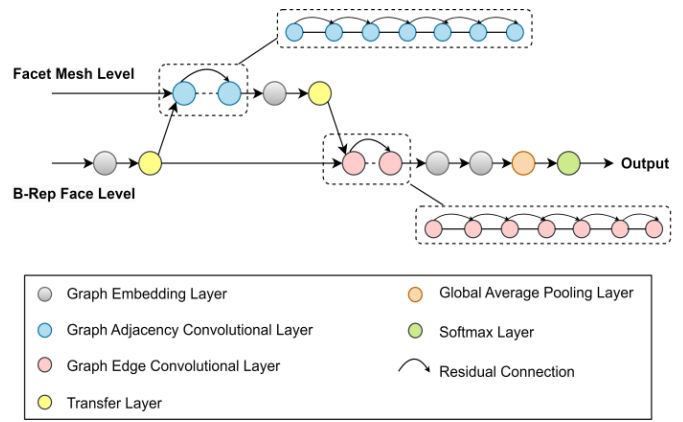

**FIGURE 5:** GRAPH NEURAL NETWORK ARCHITECTURE

The GCNN model is based on basic graph theory principles, transforming CAD models into graphs. These graphs consist of vertices (or nodes) and edges, arranged in an adjacency matrix to outline the model's structure. This setup is necessary for understanding the complex spatial relationships in CAD dataset. An important component of this architecture is the spatial graph convolution layer [24]. It is designed to work with heterogeneous graphs, improving the context of each node by gathering features from the node and its neighbors. This layer adopts a unique convolution approach, enabling the model to manage a wide variety of vertex arrangements and configurations. This capability is essential for the efficient processing of diverse CAD models.

Convolutional neural networks (CNNs) typically include pooling layers after convolutional layers to reduce the network's reliance on the exact location of features. This is achieved by downsampling and summarizing convolutional layer outputs, which reduces dimensionality. As a result, it allows for the use of larger convolutional filters, improving computational efficiency [25]. A transfer layer is then applied to facilitate the flow of information across the model's hierarchical layers. It uses linear operators (weight matrices) to facilitate the exchange of information between different dimensional features at each graph level [26]. This improves the model's understanding of the geometric and topological characteristics of the CAD model.

The softmax layer handles classification by assigning probabilities to each class in dataset. The class with the highest probability is chosen as the predicted class, corresponding to the manufacturing feature class of the B-Rep face. As the number of graph convolutional layers increases, the model initially exhibits enhanced accuracy; however, beyond a certain number of layers, a decline in performance was observed. This decline is attributed to the phenomena of vanishing gradients and oversmoothing, where the essential features become too diluted through multiple layers to effectively contribute to accurate predictions. These issues result in the backpropagated error through the network becoming very small [27]. Residual connections are employed to mitigate the vanishing gradient effect, thus ensuring the robustness of the feature recognition model. When a CAD model is input into this system, it analyzes the features and predicts their labels. The model, trained on a wide range of features in CAD models, associates them with the appropriate labels. It is, therefore, capable of accurately determining the feature labels of new CAD models.

## 2.3 Geometrical Dimensions of Recognized Features

After successfully recognizing features using the graph convolutional neural network (GCNN) model, the next step involves determining their geometrical dimensions. This involves calculating three basic geometric dimensions:
- Geometric size of a feature
- Feature orientation
- Stock sizes (minimum and maximum)

Geometric dimensions specify the unique characteristics of each feature, such as radius, depth, length, width, vertices, angle, or center. The specific dimensions to be measured depend on the type of feature. For example, for a rectangular feature, the length, width, and depth would be measured, while for a cylindrical feature, the diameter and height would be measured. These measurements are crucial for tailoring the manufacturing process to enhance adaptability across various feature types. Stock sizes include both minimum/base and maximum stock size of material. The minimum stock size represents the initial dimensions of the material with only machining features, while the maximum stock size includes all dimensions after adding both additive and subtractive features. This distinction is necessary for efficient material planning and utilization in hybrid manufacturing. Feature orientation refers to the direction of each feature's depth relative to the stock axes. It is important to determine this orientation to ensure that features are executed properly and align with the workpiece's overall design specifications. The following sections detail the extraction process for each geometric attribute.

### 2.3.1 Geometric Size Extraction of a Feature

For each recognized feature, a specialized, object-oriented PyOCC-based script is invoked. Each feature has its own script with a unique set of rules, and the appropriate script is called based on the recognized feature. These scripts interact directly with CAD models, employing PyOCC and additional computational libraries to extract the relevant feature geometrical dimensions as shown in Figure 6.

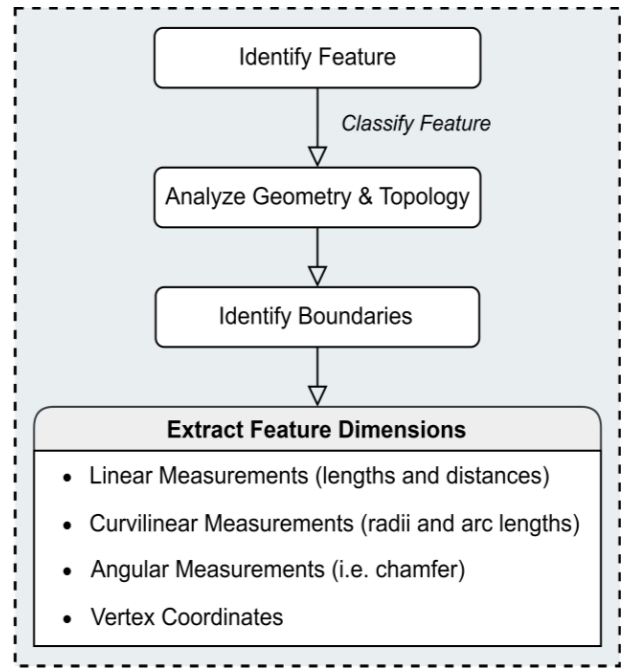

**FIGURE 6:** PROCESS FLOWCHART FOR THE EXTRACTION OF GEOMETRICAL DIMENSIONS OF A FEATURE

The process begins with the utilization of built-in Topological Exploration Tools in PyOCC [19] to analyze the CAD model's topological entities, such as edges, faces, and vertices, which define features.

These entities are then transformed into analyzable geometric objects using PyOCC built-in Geometrical Exploration Tools to extract geometric properties such as center coordinates, axes, and radii. Vector orientations are also determined during this phase by analyzing the directional attributes of these geometric shapes, crucial for defining the boundaries and orientations of features.

The next step involves the extraction of specific dimensions for each feature. This includes identifying boundaries of each feature and extracting both linear and curvilinear dimensions depending on the type of feature involved. Linear dimensions typically include lengths and distances, while curvilinear dimensions include radii, diameters, and arc lengths.

The determination of linear dimensions, such as the length (L) or distance (D), is mathematically represented as:

$$L, D = \sqrt{(x_2 - x_1)^2 + (y_2 - y_1)^2 + (z_2 - z_1)^2} \quad (1)$$

where $(x_1, y_1, z_1)$ and $(x_2, y_2, z_2)$ are the coordinates of two distinct points on the feature's boundary.

Curvilinear dimensions, such as the radius (R) of a circular feature, are derived from the parametric equation of a circle in a plane:

$$R = \sqrt{(x - h)^2 + (y - k)^2} \quad (2)$$

where $(h, k)$ are the coordinates of the circle's center, and $(x, y)$ represents any point on the circumference.

For certain features, such as chamfers, the angular dimensions also need to be calculated. This involves analyzing the geometric relationships between faces and edges, and calculating the angles at which faces intersect. The angle ($\theta$) of a chamfer or any feature is calculated based on the dot product of two vectors A and B, which correspond to the intersecting edge:

$$\cos\theta = \frac{A \cdot B}{|A||B|} \quad (3)$$

To elucidate this methodology, the example of a blind hole feature is presented. The script calculates the coordinates of the hole's top face center (C1) and the bottom face center (C2) and determines the radius (R). The depth (d) is calculated as the Euclidean distance between C1 and C2 along the hole's depth axis, represented by:

$$d = \|\overrightarrow{C2} - \overrightarrow{C1}\| \quad (4)$$

where $\overrightarrow{C2}$ and $\overrightarrow{C1}$ are the position vectors of the top and bottom face centers, respectively. Figure 7 illustrates the blind hole dimension extraction method.

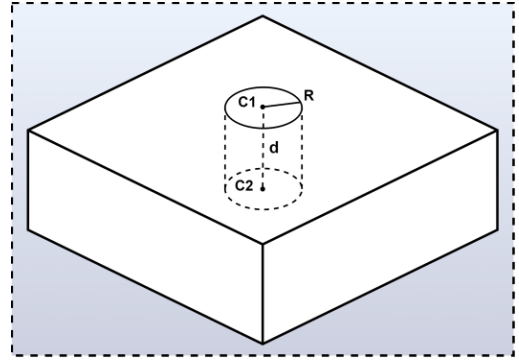

**FIGURE 7:** ILLUSTRATION OF AN EXTRACTED BLIND HOLE DIMENSIONS

To further validate the proposed method, dimension extraction of a circular end pocket feature is demonstrated. The script identifies the circular edges of the pocket, from which it calculates the centers of the end circles. The diameter (D) of the pocket is calculated as twice the radius (R), providing a measure of the pocket's overall width:

$$D = 2R = \sqrt{(x_1 - x_2)^2 + (y_1 - y_2)^2} \quad (5)$$

where $(x_1, y_1)$ and $(x_2, y_2)$ are the coordinates of diametrically opposite points on the pocket edge. The depth (d) of the pocket is measured by calculating the maximum distance between the vertices on the pocket's boundary and the plane defined by the circular edge, along the normal vector ($\vec{n}$) to this plane:

$$d = \max(|\vec{n} \cdot (\vec{v_i} - C)|) \quad (6)$$

where $\vec{v_i}$ represents the coordinates of a vertex on the pocket's boundary. The slot length (L) of the pocket is the straight-line distance between the centers of the two circular ends of the pocket. It is calculated using the Euclidean distance formula:

$$L = \sqrt{(x_2 - x_1)^2 + (y_2 - y_1)^2 + (z_2 - z_1)^2} \quad (7)$$

where $(x_1, y_1, z_1)$ and $(x_2, y_2, z_2)$ are the coordinates of the centers of circular ends of the pocket. Figure 8 illustrates the circular end pocket dimension extraction method.

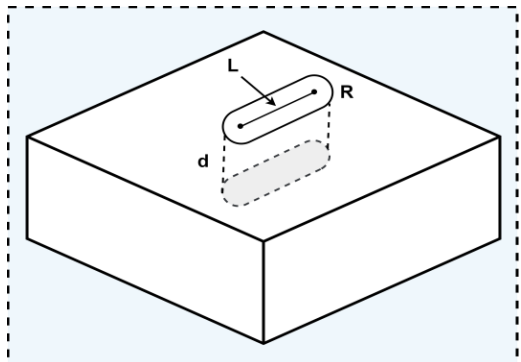

**FIGURE 8:** ILLUSTRATION OF AN EXTRACTED CIRCULAR END POCKET DIMENSIONS

**2.3.2 Feature Orientation Extraction**

After the extraction of geometrical dimensions, the orientation of recognized features within the CAD model is determined through geometrical analysis. Features are classified based on their orientation into two categories: upright and tilted.

Upright features have normal vectors that align with the global Z-axis, indicating perpendicularity to the model's base plane. This alignment simplifies machining operations due to straightforward tool access and minimal angular adjustments. Conversely, tilted features have normal vectors aligned along the X or Y axis, showing a significant angular deviation from the Z-axis. Such features are not perpendicular to the base plane, requiring specialized machining strategies for their unique orientation. For instance, a cylindrical hole is considered upright if its depth axis (the direction along which the normal vector is calculated) aligns with the Z-axis, and tilted if it aligns more with the X or Y axes. Figure 9 shows a visual comparison of upright and tilted features.

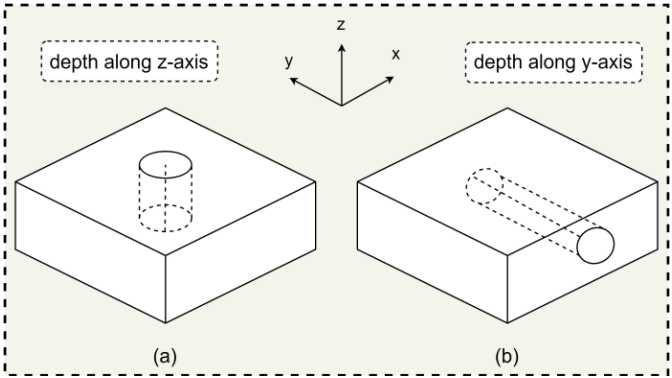

**FIGURE 9:** COMPARISON OF (a) UPRIGHT AND (b) TILTED FEATURE

The process begins by calculating the normal vector for each feature's primary surface, crucial for determining its orientation relative to the global coordinate system. This is achieved by analyzing geometric data from the CAD model, including the positions of vertices and the orientations of faces. For instance, in the case of a hole, the normal vector is derived from its defining axis. This method is applied selectively to features based on their specific geometrical configurations, which make the analysis relevant, and can accurately represent the feature's orientation in three-dimensional space through mathematical operations.

After computing the normal vectors, the script compares them to the global Z-axis for upright features, and to the X or Y axes for tilted features. This comparison assesses the angular deviation using trigonometric functions to measure the angle between the feature's normal vector and these axes. If the angle indicates alignment with the Z-axis, the feature is classified as upright. If it aligns more closely with the X or Y axes, it is classified as tilted.

This classification into upright and tilted features is pivotal for developing subsequent manufacturing strategies, guiding the selection of tools and paths that align with the orientation of each feature.

### 2.3.3 Stock Size Calculation

Building upon the extracted feature geometrical dimensions and orientation analysis, stock sizes are calculated for manufacturing process decision-making. The stock sizes are categorized into two types: minimum stock size and maximum stock size.

The calculation of the minimum stock size focuses on the essential volume needed for machining by employing a bounding box analysis. This analysis identifies the smallest enclosing box around all necessary machining features, excluding additive features to concentrate solely on the material required for traditional machining. Spatial transformations such as translation, rotation, and scaling are applied dynamically to adjust each feature within the bounding box, ensuring an accurate determination of the minimum material volume required. These transformations help align and size all the features accurately within the bounding box:

$$\text{Translation:} \quad \vec{p}' = s'\vec{p} \quad (8)$$
$$\text{Rotation:} \quad \vec{p}' = R\vec{p} \quad (9)$$
$$\text{Scaling:} \quad \vec{p}' = s\vec{p} \quad (10)$$

where $\vec{p}$ and $\vec{p}'$ are the original and transformed points, respectively, $s'$ is the translation vector, $R$ is the rotation matrix, and $s$ is the scaling factor. Mathematically, for a bounding box defined by its maximum extents $x_{max}$, $y_{max}$, $z_{max}$ and minimum extents $x_{min}$, $y_{min}$, $z_{min}$, its dimensions are calculated as:

$$x_{size} = x_{max} - x_{min} \quad (11)$$
$$y_{size} = y_{max} - y_{min} \quad (12)$$
$$z_{size} = z_{max} - z_{min} \quad (13)$$

where $x_{size}$, $y_{size}$ and $z_{size}$ represent the dimensions of the model's bounding box in the x, y, and z directions, respectively.

For the maximum stock size, the same spatial transformations are applied to encompass both additive and machining features, ensuring the entire volume needed for manufacturing is accurately identified. The global bounding box for the CAD model includes all features, accommodating additional volumes required for additive features. This comprehensive approach ensures that the calculated stock size is precisely tailored to the part's specifications, considering its geometric complexities. Figure 10 clearly shows the minimum and maximum stock sizes, calculated using the bounding box method. This illustration demonstrates how both stock sizes effectively encompass all the features within the base cuboidal stock.

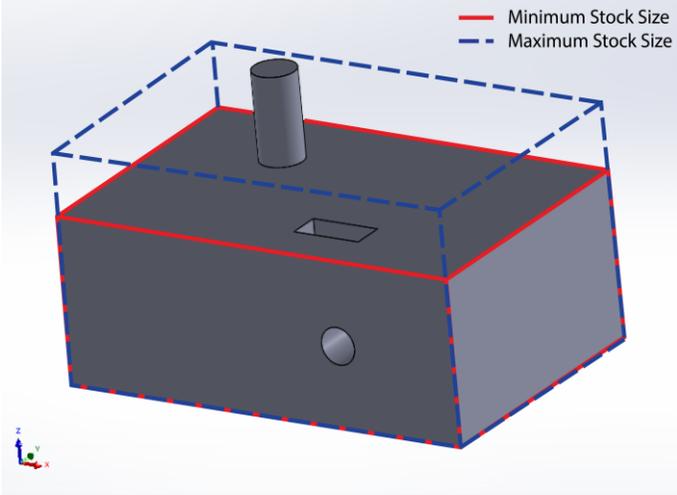

**FIGURE 10:** A CAD MODEL WITH EXTRACTED STOCK SIZES

## 3 RESULTS AND DISCUSSION

This study presents a novel method for recognizing hybrid manufacturing features and extracting dimensional attributes in CAD models. The proposed method bridges the gap between CAD designs and manufacturing processes by accurately identifying and characterizing features, including their dimensions, orientations, and stock sizes.

A hybrid feature recognition model is developed using TensorFlow 2 and executed on an NVIDIA GeForce RTX 3090 GPU. The network's weights are initialized using the Xavier initialization method [28], and it includes seven graph convolutional layers at each level of the GCNN model. Each convolutional layer is followed by batch normalization [29] and dropout [30]. Dropout rate is maintained at 0.3 to prevent overfitting. The model employs 512 filters per graph convolutional layer and utilizes a learning rate that decays from an initial value of 0.01. The ADAM optimizer [31] is used for optimization, and categorical cross-entropy [32] is employed for loss calculation. Each training batch consists of a variable number of hierarchical graphs, with the total number of vertices across all graphs in the batch being less than 5000. The dataset is split 60:20:20 for training, validation, and testing across 100 epochs. The dataset consists of 150,000 synthetic labeled CAD models with 29 distinct manufacturing feature classes. Each CAD model contains 4 to 8 manufacturing features. The evaluation metrics for the proposed model include accuracy, precision, recall, and the F1 score, calculated as follows: [33]

$$Accuracy = \frac{TP + TN}{TP + TN + FP + FN} \quad (14)$$

$$Precision = \frac{TP}{TP + FP} \quad (15)$$

$$Recall = \frac{TP}{TP + FN} \quad (16)$$

$$F1\ score = 2 \times \frac{Precision \times Recall}{Precision + Recall} \quad (17)$$

where TP, TN, FP, and FN are true positive, true negative, false positive, and false negative, respectively. The results of the model evaluation metrics are shown in Table 1.

**TABLE 1:** COMPREHENSIVE EVALUATION METRICS FOR THE FEATURE RECOGNITION MODEL

| Evaluation Metric | Value (%) |
| --- | --- |
| Accuracy | 96.87 |
| Precision | 97.16 |
| Recall (Sensitivity) | 96.67 |
| F1 score | 96.87 |

This section presents a comparative analysis of a CAD model with various manufacturing features against the same processed model, where the extracted features are highlighted in different colors. The proposed model processes each B-Rep face of the cuboidal stock to extract features, including dimensions, orientations, and stock sizes. This detailed analysis is essential for validating the predictive accuracy of our model.

Figure 11 shows a CAD model with hybrid features, serving a baseline for the proposed feature extraction analysis. After being processed through the neural network model, the features are accurately recognized and their geometrical dimensions, along with orientation and stock sizes, are precisely extracted. Figure 12 illustrates the CAD model post-analysis, with features distinctly highlighted according to the model's predictions. Table 2 complements the analysis by detailing the specific features extracted by the proposed model. It enumerates each feature's type and index, alongside their extracted dimensions and orientations, providing a clear view of the model's accuracy in identifying and characterizing manufacturing features within CAD models.

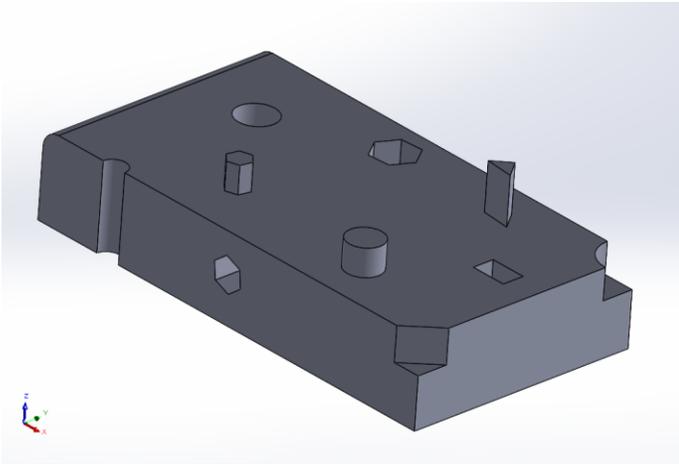

**FIGURE 11:** A TEST CAD MODEL WITH HYBRID FEATURES

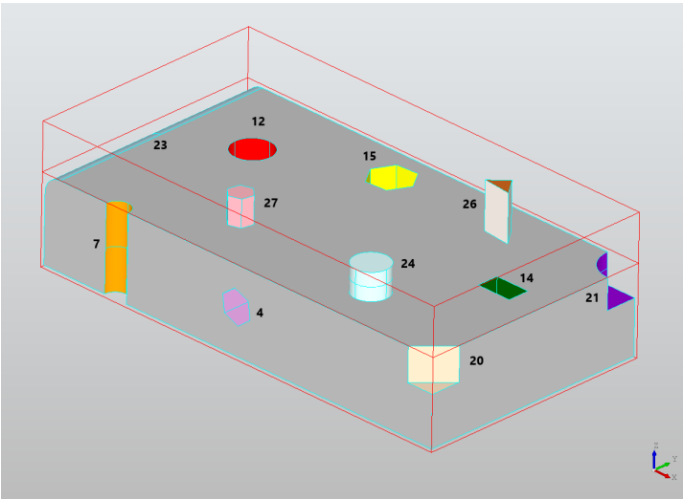

**FIGURE 12:** CAD MODEL WITH RECOGNIZED FEATURE LABEL AND COLOR AFTER FEATURE RECOGNITION

**TABLE 2:** DETAILS OF FEATURES IDENTIFIED BY DIMENSION EXTRACTION METHOD

| Feature Type with Index | Extracted Dimensions | Feature Orientation |
|---|---|---|
| Blind Hole (12) | Radius = 5.92, Depth = 25.0, Center = (-40.24, 10.48, 25.0) | Upright |
| 6-sided Pocket (15) | Side = 6.77, Depth = 25.0 | Upright |
| Hexagonal Extrusion (27) | Side = 3.67, Depth = 8.0 | Upright |
| Cylindrical Extrusion (24) | Radius = 5.3, Depth = 8.0, Center = (31.23, -21.23, 33.0) | Upright |
| Triangular Extrusion (26) | Side-1 = 6.77, Side-2 = 6.47, Side-3 = 8.7, Depth = 14.0 | Upright |
| Rectangular Pocket (14) | Length = 6.43, Width = 10.48, Depth =17.0 | Upright |
| 6-sided Passage (4) | Side = 5.05, Depth = 22.0 | Tilted |
| Circular Blind Step (21) | Radius = 9.47, Depth = 9.0, Centre = (65.9, 34.84, 16.0) | Upright |
| Triangular Blind Step (20) | Length = 9.18, Width = 8.39, Side = 12.44, Depth = 10.0 | Upright |
| Round (23) | Radius = 5.0, Face index (6, 6) | None |
| Circular Thru Slot (7) | Radius = 3.75, Depth = 25.0, Center = (-41.49, -34.84, 25.0) | Upright |

The minimum and maximum stock sizes for the entire test model are [131.81 × 69.69 × 25] and [131.81 × 69.69 × 39], respectively. For the test CAD model discussed, all features and dimensions have been determined with 100% accuracy. The accuracy of feature recognition is critical as it directly influences the efficiency of the manufacturing process. Correctly identified feature dimensions guide tool paths and equipment setup, which can significantly reduce manufacturing time and material waste. Furthermore, understanding the orientation of features is essential for ensuring that parts are manufactured correctly and fit their intended use.

## 4 CONCLUSIONS AND FUTURE WORK

This paper introduced a novel approach for enhancing manufacturing decision-making by recognizing hybrid additive-subtractive manufacturing features and extracting geometric attributes in CAD models. The proposed method precisely identified, characterized, and quantified manufacturing features, effectively bridging the gap between CAD designs and manufacturing processes. By leveraging a comprehensive dataset of CAD models with hybrid features, the graph convolutional neural network model achieved a feature recognition accuracy of around 97%.

However, there are limitations in this study. The accurate extraction of dimensional attributes is inherently dependent on the correct recognition of feature labels. An incorrect feature recognition leads to imprecise dimensional extraction, as the process relies on accurate initial feature identification. Furthermore, the dimension extraction process for intersecting features poses additional challenges, as it tends to analyze and extract dimensions of intersected features in sections, rather than as a unified feature.

For future work, we aim to refine the accuracy of recognizing and extracting dimensions for intersecting features, alongside enhancing the model to identify a broader range of machining and additive manufacturing features. Additionally, we plan to extend the proposed model's capabilities to calculate more detailed parameters such as surface area, normal vectors, and volume for each feature. Furthermore, we will combine the AFR model with a manufacturing decision-making system, creating an enhanced ecosystem that supports more informed and efficient manufacturing processes.

**ACKNOWLEDGEMENTS**

This work is supported by the Agency for Science, Technology, and Research (A*STAR), Singapore through the RIE2025 MTC IAF-PP grant (Grant No. M22K5a0045). It is also supported by Singapore International Graduate Award (SINGA) (Awardee: Muhammad Tayyab Khan) funded by A*STAR and Nanyang Technological University, Singapore.